\pdfoutput=1

\documentclass[11pt]{article}

\usepackage[final]{acl}

\usepackage{times}
\usepackage{latexsym}

\usepackage[T1]{fontenc}

\usepackage[utf8]{inputenc}

\usepackage{microtype}

\usepackage{inconsolata}

\usepackage{graphicx}

\usepackage{amssymb}
\usepackage{algorithm}
\usepackage{algpseudocode}
\usepackage{tabularx}
\usepackage{caption}
\usepackage{subcaption}
\usepackage{rotating}
\usepackage{array}
\usepackage{xspace}
\usepackage{bbm}
\usepackage{scalerel}
\usepackage{microtype}
\usepackage{color}
\usepackage{multirow}
\usepackage{graphicx}
\usepackage{booktabs}
\usepackage{bbding}
\usepackage{makecell}
\usepackage{amsmath}
\usepackage{natbib}
\usepackage{ifthen}
\usepackage{enumerate}
\usepackage{float} 
\usepackage{colortbl} 
\usepackage{blindtext} 
\usepackage{xcolor, soul} 
\usepackage[super]{nth} 
\usepackage[capitalize]{cleveref}
\usepackage{bookmark}

\usepackage{tcolorbox}
\usepackage{pifont}       
\usepackage{bbding}       
\usepackage{fontawesome}
\usepackage{soul}
\usepackage{multirow}

\title{Verifiable Format Control for Large Language Model Generations}

\makeatletter
\def\@fnsymbol#1{\ensuremath{\ifcase#1\or \dagger\or *\or \ddagger\or
   \mathsection\or \mathparagraph\or \|\or **\or \dagger\dagger
   \or \ddagger\ddagger \else\@ctrerr\fi}}
\makeatother

\def\myand{\end{tabular}\hss\egroup \hfil\hfil\egroup
           \hbox to \linewidth\bgroup\large \hfil\hfil
             \hbox to 0pt\bgroup\hss \begin{tabular}[t]{c}\bf}

\author{
    Zhaoyang Wang\thanks{~Equal contribution.}$\,^{1}$~~~Jinqi Jiang\footnotemark[1]$\,^{1}$~~~Huichi Zhou\footnotemark[1]$\,^{2}$\\
    \textbf{~Wenhao Zheng}$^1$~~~\textbf{Xuchao Zhang}$^3$~~~\textbf{Chetan Bansal}$^3$~~~\textbf{Huaxiu Yao}$^{1}$\\
    $^1$University of North Carolina at Chapel Hill\\
    $^2$Imperial College London\quad$^3$Microsoft Research\\
    {\texttt \{zhaoyang,~huaxiu\}@cs.unc.edu}
}

\newcommand{\dataset}{\texttt{VFF}\xspace}

\begin{document}
\maketitle
\begin{abstract}

    Recent Large Language Models (LLMs) have demonstrated satisfying general instruction following ability. However, small LLMs with about 7B parameters still struggle fine-grained format following (e.g., JSON format), which seriously hinder the advancements of their applications. Most existing methods focus on benchmarking general instruction following while overlook how to improve the specific format following ability for small LLMs. Besides, these methods often rely on evaluations based on advanced LLMs (e.g., GPT-4), which can introduce the intrinsic bias of LLMs and be costly due to the API calls. In this paper, we first curate a fully verifiable format following dataset \dataset. In contrast to existing works often adopting external LLMs for instruction-following validations, every sample of \dataset can be easily validated with a Python function. Further, we propose to leverage this verifiable feature to synthesize massive data for progressively training small LLMs, in order to improve their format following abilities. Experimental results highlight the prevalent limitations in the format following capabilities of 7B level open-source LLMs and demonstrate the effectiveness of our method in enhancing this essential ability.

\end{abstract}

\section{Introduction}

Recent advancements in Large Language Models (LLMs) have demonstrated a series of foundation abilities including in-context learning, reasoning, and the essential instruction following ability~\citep{brown2020language,wei2022emergentabilitieslargelanguage,openai2023gpt4,bubeck2023sparks}.
Pre-trained LLMs such as GPT-3~\citep{brown2020language} hardly follow human instructions due to the mismatch issue between their pre-training objectives and human preferences~\citep{zhang2023instruction}. 
To address this issue, a series of works employ instruction tuning to enable LLMs to respond fluently to natural questions~\citep{longpre2023flan,touvron2023llama,xu2023wizardlm}, which effectively align LLMs with human preferences. 
Specifically, these methods may first collect instruction-response pairs from human~\citep{vicuna2023,zhou2023lima,mishra2021cross} or more powerful LLMs~\citep{alpaca,wang-etal-2023-self-instruct,xu2023wizardlm} (e.g., ChatGPT~\citep{ouyang2022training}). Then, these collected data can be used to fine-tune LLMs to follow human desired responses. 
Further, \citet{ouyang2022training} propose Reinforcement Learning from Human Feedback (RLHF) to enhance the alignment of LLMs, improving the helpfulness and harmfulness of the generations~\citep{bai2022training}.
Today's advanced LLMs such as GPT-4~\citep{openai2023gpt4} can follow most human instructions even those with fine-grained format control requirements (e.g., JSON output).
However, more widely used open-source 7B-level LLMs such as Mistral~\citep{jiang2023mistral} and LLaMA series~\citep{llama2,llama3} often struggle with fine-grained format control despite achieving satisfactory results in general instruction following.
In this paper, we focus on enhancing such fine-grained format control ability of small LLMs to benefit LLM-based applications especially for the format-sensitive ones.

\paragraph{Evaluating.} First, we propose to evaluate the fine-grained format control ability of LLMs.
Most existing research~\citep{qin2024infobench,ifeval,jiang2023followbench, yizhi2024cifbench,ma2024activate} in this area proposes general instruction following benchmarks, while paying less attention to specific format control.
Also, in terms of verifying and evaluation, most of them are driven by LLMs based evaluations which heavily rely on the capability of the selected LLMs~\citep{chiang-lee-2023-large,fu2023gptscore,liu2023calibrating,chan2023chateval}. Further, few of them consider improving the format following ability of small LLMs.

\begin{figure}[t]
\centering
\includegraphics[width=\columnwidth]{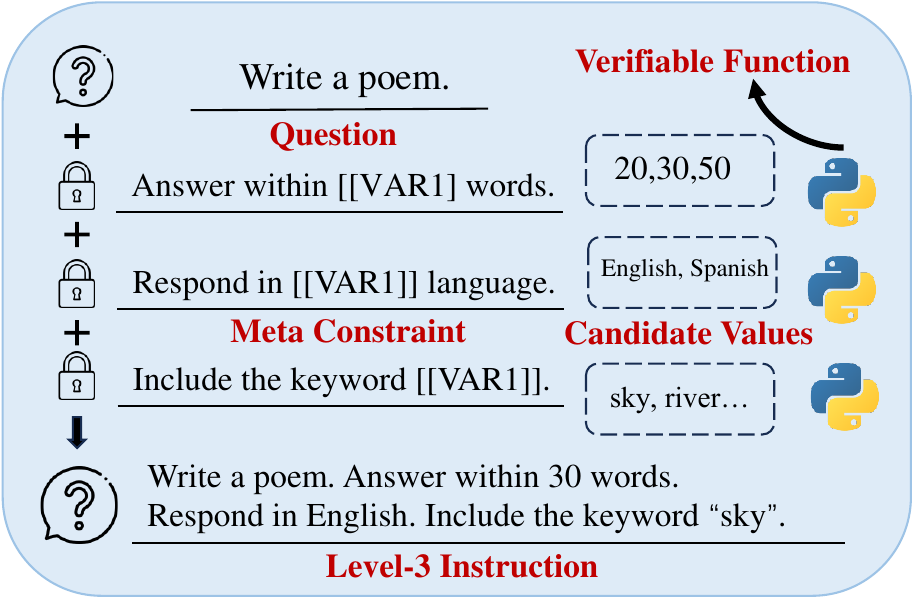}
\caption{An example with level-3 constraints of \dataset.}
\label{fig:data-instance}
\end{figure}

To address these challenges, we curate a fully \textbf{V}erifiable \textbf{F}ormat \textbf{F}ollowing (\dataset) dataset. 
This dataset starts with a few GPT-4 annotated meta constraints. And we can use off-the-shelf Alpaca dataset~\citep{alpaca} as the prompt (question) source.
As illustrated in Figure~\ref{fig:data-instance}, the prompt combined with the meta constraints can finally form an instruction with various format controls.
Specifically, each meta constraint consists of several variables and candidate values (which formulates the constraint instance), and a corresponding Python function (which verifies the format following).
Note that the final instantiated  instruction can include multiple different constraints, referred to as multi-level constraints. These multi-level constraints can go up to $3$ levels, considering the needs in realistic scenarios.
We also detail the differences between our \dataset and existing datasets in Table~\ref{tab:benchmark_comparison}.

\begin{table}[t]
\small
\centering
\caption{Comparison with existing benchmarks. ``ML'' refers to multi-level constraints. ``HJ'', ``LJ'', ``AJ'' indicate the judgments relying on human, LLMs and automatic scripts (e.g, Python), respectively.}

\resizebox{\linewidth}{!}{
\begin{tabular}{l|cc|ccc}
\toprule
 Benchmark      & Verifiable & ML & HJ & LJ & AJ  \\ \midrule
Alpaca~\citep{alpaca} & \ding{56}  & \ding{56}   & \ding{52} & \ding{56} & \ding{56}              \\ 
Vicuna~\citep{vicuna2023} & \ding{56}  & \ding{56}   & \ding{52} & \ding{56} & \ding{56}     \\ 
PandaLM~\citep{wang2023pandalm} & \ding{56}  & \ding{56}   & \ding{52} & \ding{56} & \ding{56}           \\
Wizardlm~\citep{xu2023wizardlm} & \ding{56}  & \ding{56}   & \ding{52} & \ding{56} & \ding{56}             \\
LLM-EVAL~\citep{lin-chen-2023-llm} & \ding{56}  & \ding{56}   & \ding{56} & \ding{52} & \ding{56}              \\
IFEval~\citep{ifeval} & \ding{52}  & \ding{52}   & \ding{56} & \ding{56} & \ding{52}                 \\ 
FollowBench~\citep{jiang2023followbench} & \ding{56}  & \ding{52}   & \ding{56} & \ding{52} & \ding{52}         \\
INFOBENCH~\citep{qin2024infobench} & \ding{56}  & \ding{56}   & \ding{56} & \ding{52} & \ding{56}                 \\
FCS~\citep{he2024complex} & \ding{56}  & \ding{52}   & \ding{56} & \ding{52} & \ding{56}                \\
Conifer~\citep{sun2024conifer} & \ding{56}  & \ding{56}   & \ding{56} & \ding{52} & \ding{56}                \\
FOFO~\citep{xia2024fofo} & \ding{56}  & \ding{56}   & \ding{56} & \ding{52} & \ding{56}             \\
\midrule
\dataset (ours)  & \ding{52}  & \ding{52}   & \ding{56} & \ding{56} & \ding{52}                \\ \bottomrule
\end{tabular}}
\label{tab:benchmark_comparison}
\end{table}

\paragraph{Enhancing.} After having \dataset to evaluate the format following ability, we then propose to leverage \dataset's easily verifiable feature to improve such important abilities for 7B level LLMs.
Previous methods~\citep{vicuna2023,xu2023wizardlm,wang-etal-2023-self-instruct} often do not specially introduce fine-grained format following data, leading to inferior performance in this area.
Besides, the training data used by these methods are often collected via human sharing like ShareGPT\footnote{\url{https://sharegpt.com/}} or advanced LLMs like GPT-4, which can be costly and time-consuming.
Thanks to the easily verifiable feature of our \dataset, we can synthesize training data to improve the format following ability of LLMs in a self-improvement paradigm.
This paradigm ensures the training data is entirely generated by the LLM itself.
Specifically, this paradigm consisted of three stages: (1) Sampling multiple responses from the LLM for each instruction with constraints from \dataset. (2) Annotating the responses using the verifiable Python function to identify whether they strictly follow the format controls.  (3) Fine-tuning the LLM via supervised fine-tuning and preference learning (e.g., DPO~\citep{rafailov2024direct}) with those annotated responses.
Furthermore, these steps can be repeated in a progressive training manner,  starting by training the LLM to follow a single constraint (level-1) and advancing to adhere to multiple constraints (level-3).

In summary, our contributions are as follows:
\begin{enumerate}[1)]
    \item We curate a fully verifiable format following dataset \dataset, showing that 7B level LLMs hold potential for enhanced format control ability.
    \item With verifiable feature of \dataset, we propose the progressive training to enhance LLMs' format control ability with  self-generated training data.
    \item Empirical results on existing benchmarks with several trained 7B-level LLMs demonstrate the effectiveness of the proposed method.
\end{enumerate}

\section{Related Work}

\subsection{Evaluating Instruction Following} 

To evaluate the instruction following capabilities of large language models (LLMs), three primary methods are commonly employed: human evaluation, LLM-based evaluation, and evaluation through verifiable instructions (i.e., automatic evaluation). While human evaluation can be accurate, it suffers from subjective bias, high costs, and a lack of reproducibility~\citep{ouyang2022training, bang2023multitask, wang2023pandalm, xu2023wizardlm}. 
LLM-based evaluation methods offer more scalable and robust alternatives for assessing instruction following performance~\citep{lin-chen-2023-llm, liu2303g, qin2024infobench, he2024complex, sun2024conifer, xia2024fofo}. 
For automatic evaluations, \citet{jain2023bring} analyze the sensitivity of the slight changes in LLM outputs as a means of measuring their reliability, while \citet{ifeval} introduce the IFEval benchmark, which directly verifies LLMs' instruction-following abilities through simple code execution. Similarly, FollowBench~\citep{jiang2023followbench} proposes a benchmark that integrates LLMs with Python scripts to act as an evaluator.

First, this paper focuses on evaluating the format following ability of LLMs instead of verifying the following of the instruction. For example, checking the hallucinations of the response content is beyond our research focus.
Next, to effectively evaluate the format following ability, we propose \dataset dataset which uses automatic evaluation through Python functions.
Compared to human or LLM-based evaluations, our method can provide a more reliable and efficient way to examine whether LLMs adhere to specific format controls.
In addition to existing verifiable benchmarks, our dataset covers more types of format constraints such as JSON output. Further, we propose a training pipeline to enhance LLM's format control ability with such verifiable feature. 

\subsection{Enhancing Instruction Following}
InstructGPT~\citep{ouyang2022training}  proposes RLHF to train the GPT-3 model~\citep{brown2020language} to follow human instructions.
Subsequent research has focused on developing open-domain general instruction following datasets including Alpaca~\citep{alpaca} and Vicuna~\citep{vicuna2023}, both of which played a key role in enabling LLaMA~\citep{touvron2023llama} to follow instructions.
Additionally, WizardLM~\citep{xu2023wizardlm} utilizes AI-generated data for instruction fine-tuning, offering control over the complexity and difficulty of the instructions.
Following the introduction of Direct Preference Optimization (DPO)~\citep{rafailov2024direct}, a number of recent works~\citep{jiang2023preference, ethayarajh2024kto, hong2024orpo, meng2024simpo} have proposed preference learning algorithms aimed at enhancing the instruction following and alignment performance of LLMs.
Another line of work explores constrained decoding methods to enhance the structured generation performance~\citep{openai2023structured,outlines2023,lmql2023,jsonformer2023,dong2024xgrammar}. However, the range of supported structured formats may be limited.

In this paper, our goal is to improve the format following ability of widely used 7B-level LLMs, rather than focusing on general instruction following. 
By leveraging self-generated training data, we propose a progressive training approach that iteratively trains small LLMs to follow format constraints of varying difficulty levels. 
An additional benefit of this method is its scalability, as the training data can be easily synthesized due to the verifiable nature of our pre-curated \dataset dataset.

\section{Method}

In this section, we first introduce the curation of our verifiable format following dataset \dataset, which is for accurately evaluating the format control ability of LLM generations.
By leveraging the easily verifiable feature of \dataset, we then propose a progressive training manner to  enhance such format control ability of 7B level LLMs.

\subsection{\dataset dataset}

\begin{figure}[t]
\centering
\includegraphics[width=\columnwidth]{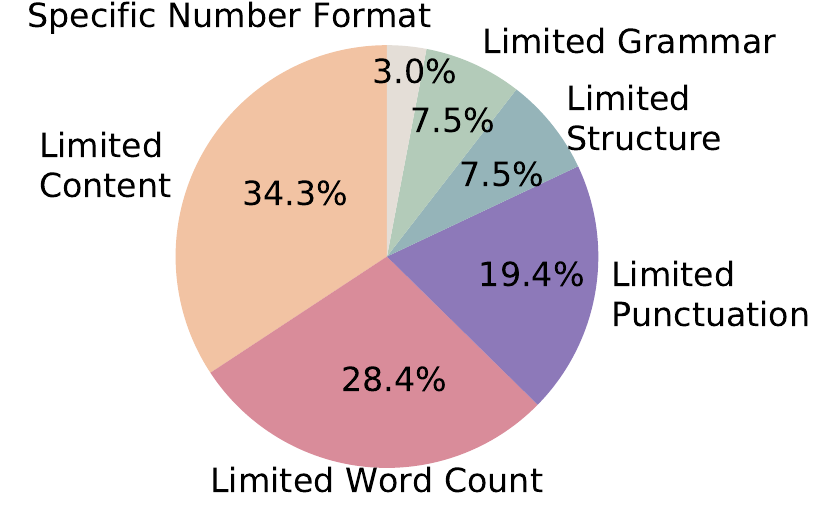}
\caption{Category distribution of meta constraints. The ``specific number format'' often involves constraints on the generated numbers (e.g., time format). The ``limited grammar'' includes constraints for writing styles such as active or passive voice. The ``limited structure'' includes common structure output formats such as JSON, YAML and etc. The ``limited punctuation'' requires the LLMs to use specific punctuations in the generations. The ``limited word count'' directly limits the length of the output from LLMs.  The ``limited content'' constrains generations within specific topics, which can limit the output scope of the response.}
\label{fig:dist-vff}
\end{figure}

\begin{figure*}[t]
\centering
\includegraphics[width=0.98\textwidth]{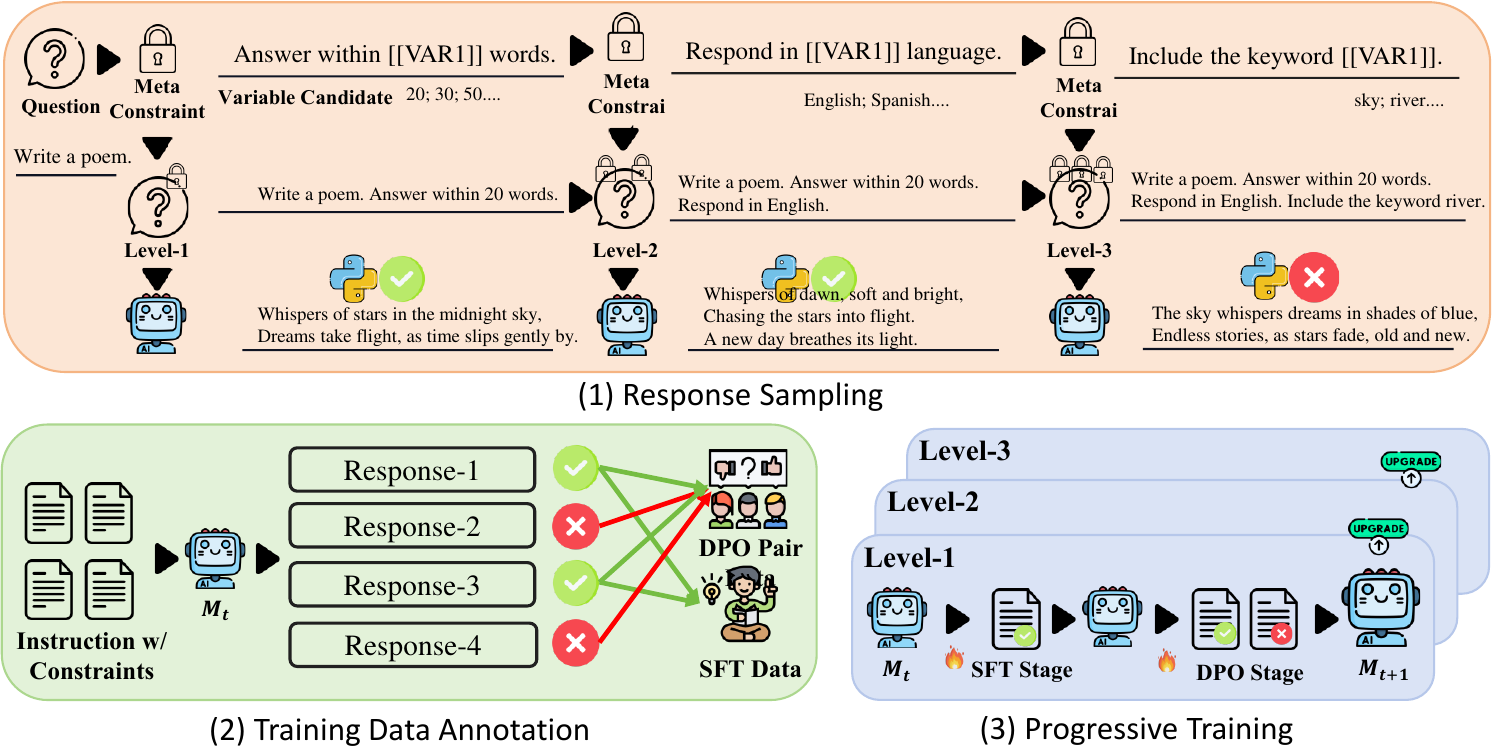}
\caption{The pipeline of the proposed method for enhancing format control ability. First, the model takes response sampling for instantiated instructions from \dataset dataset. Next, in the training data annotation stage, by utilizing the verifiable Python functions, we can collect the responses that satisfy the constraints as the SFT data, while pair with the negative responses to form the DPO training data.
Finally, the LLM is first trained with SFT, followed by direct preference learning (DPO) to improve its format following.
These steps are repeated in a progressive training manner with the increased levels of difficulty, in order to exploit the potential of the LLM.
}
\label{fig:train-pipeline}
\end{figure*}

\paragraph{Verifiable Meta Constraint.} 
We begin with a small set of human-annotated meta constraint pool $\mathcal{D}_{M} = \{(C_k, V_k, F_k) \mid 1 \leq k \leq 16\}$, where $C_k$ represents the $k$-th constraint containing variables, $V_k$ is the set of candidate values that can be selected to fill in the variables, $F_k$ is the corresponding Python bool function that can efficiently and accurately verify whether the generated responses satisfy the format constraints.
For example, $C_k$ is ``Respond in [[VAR1]] language.'', $V_k$ is a set of values ``\{English, Spanish, French, Chinese, Japanese\}'' for ``[[VAR1]]'' to fill in, and $F_k$ is an executable Python bool function for language detection.
We extend this pool $\mathcal{D}_{M}$ by leveraging the in-context learning ability of GPT-4~\citep{openai2023gpt4}. 
In evaluating the format following scenario, the adopted Python function serving as the judgment method is as accurate as human judgment, while being more efficient and time-saving compared to LLM-based evaluations. These advantages make it highly suitable for large-scale verifications. 

Some constraints may seriously conflict with user's instruction, for example, the instruction is writing a long story, while the constraint may limit the number of generated words to $10$.
Thus, we also take manual check for each generated sample. 
Finally, to maximize the applicability and universality of our meta constraints, we only reserve about $60$ unique meta constraints, which are publicly available at \url{https://huggingface.co/datasets/jinqij/VFF}. We visualize the categories of these constraints in Figure~\ref{fig:dist-vff}. These categories can cover common realistic output format needs for LLM based applications.

\paragraph{Format Following Dataset.}
After obtaining the meta constraints, we then need to instantiate the instruction with specific format controls. These instantiated instructions form the format following dataset that can be directly used to evaluate the performance of LLMs.
Specifically, each instruction sample $x$ of \dataset dataset $\mathcal{D}_{V}$ consists of a detailed question coupled with several constraint instances. 
In details, we use the existing Alpaca dataset~\citep{alpaca} as the question source $Q$ which comprises about $52$K questions generated by text-davinci-003~\citep{brown2020language}.
Then, considering the risks in conflicts between questions and constraints, we randomly select up to $3$ ($1 \le c \le 3$) unique meta constraints from $\mathcal{D}_M$, and instantiate them by filing the variables of the constraints $C_k$ with their respective candidate values $V_k$.
The instruction sample $x$ illustrated in Figure~\ref{fig:data-instance} can be obtained by concatenating the question $q$ and the constraint instance $d$, i.e., $x = [q, d]$ where $d$ includes $c$ unique constraint instances, reflecting a $\text{level-}c$ difficulty as categorized in this study.
Simultaneously, the corresponding binary functions $F_{k \le c}(\ast)$ can serve together to verify the correctness of the generated response $y$ with $I = \prod_{k=1}^{c} F_k(y)$, where $I = 1$ represents as the model can fully adhere to this format control instruction $x$. 
Note that the size of $\mathcal{D}_{V}$ can be as large as $|Q| \times (\sum_k^{|\mathcal{D}_M|} |V_k|)$, which is considerably larger than previous benchmarks. This can be viewed as an advantage of our dataset especially in the era of scaling synthetic data.

\begin{algorithm}[t!]
    \small
    \centering
    \caption{Enhancing Format Control.}
    \begin{algorithmic}[1]
    \Require Pre-trained model $M$, format following dataset $\mathcal{D}_{V}$, verifiable functions $\{F_k(\ast)\}$
    \State \textbf{Stage 1: Response Sampling}
    \For{each instruction $x \in \mathcal{D}_{V}$}
        \State Sample $4$ responses $\{y_i\}$ from $M$ using $x$
    \EndFor
    \State
    \State \textbf{Stage 2: Training Data Annotation}
    \For{each response $y_i$}
        \State Compute $I(y_i) = \prod_{k=1}^{c} F_k(y_i)$
        \If{$I(y_i) = 1$}
            \State Mark $y_i$ as preferred response $y_w$
        \Else
            \State Mark $y_i$ as dis-preferred response $y_l$
        \EndIf
    \EndFor
    \State Obtained the training data $\mathcal{D}=\{x, y_w, y_l\}$
    \State
    \State \textbf{Stage 3: Progressive Training}
    \For{difficulty level $c$ from $1$ to $3$}
            \State Fine-tune model $M$ on $\mathcal{D}$ via Eq.~\ref{eq:sft}
            \State Fine-tune model $M$ on $\mathcal{D}$ via Eq.~\ref{eq:dpo} 
            \State Re-sample and re-annotate with fine-tuned model $M$, updating $\mathcal{D}$
    \EndFor
    \end{algorithmic}
    \label{alg:enhance}
\end{algorithm}

\subsection{Enhancing Format Control}
As shown in Figure~\ref{fig:train-pipeline}, the proposed method for enhancing the format control ability of small LLMs mainly consists of three stages: (1) Response Sampling stage, (2) Training Data Annotation stage, and (3) Progressive Training stage.

\paragraph{Response Sampling.}
To collect diverse enough responses, we instruct the LLM to sample multiple responses for the same question with constraints.
Specifically, for each instruction $x \in \mathcal{D}_V$, we sample $k=4$ responses from the LLM. However, given the limitations of 7B-level LLMs, these may all be incorrect. 
To improve sampling efficiency for correct responses, inspired by recent self-improvement studies~\citep{huang-etal-2023-large,wang-etal-2023-democratizing,gou2024critic}, we add one generated wrong response as the one-shot demonstration to help the LLM to generate better response.

\paragraph{Training Data Annotation.}
To identify the correctness of the massively generated responses, we propose to leverage the fully verifiable feature of \dataset dataset.
Specifically, the collected response samples can be efficiently annotated with format following judgments by the verifiable functions $I$, which reduces the costs of calling GPT-4 APIs compared to previous methods.
Through these verifiable functions, we can identify the response that satisfies the format constraints ($I=1$) as preferred response $y_w$, and responses that not following the constraints ($I=0$) as dis-preferred response $y_l$.
Finally, we can use the preferred responses to form training data $\mathcal{D} = \{x, y_w, y_l\}$.
In this dataset, the preferred responses $\{y_w\}$ can be used for SFT training, while the preference pairs $\{y_w, y_l\}$ can be used for DPO training. 
Note that this training data is fully generated and annotated by LLM itself, without any needs for human or external LLMs.

\paragraph{Progressive Training.}
The small LLM's fine-grained format following ability can be enhanced by aligning it with human desired output format.
Specifically, we first apply Supervised Fine-Tuning (SFT) to train the LLM $\pi$ on the self-generated good responses $y_w$, aimed at improving basic capability of format following.
The SFT training objective is detailed as follows:
\begin{equation}
    \mathcal{L}_\text{SFT} = - \mathbb{E}_{(x, y_w, y_l) \sim \mathcal{D}} \log \pi(y_w|x) .
    \label{eq:sft}
\end{equation}
With recent advancements of preferece learning methods~\citep{dong2023raft,rafailov2024direct,ethayarajh2024kto}, we can apply Direct Preference Optimization~\citep{rafailov2024direct} (DPO) to align the fine-tuned LLM more closely with the desired response formats. The DPO training objective can be formulated as follows:
\begin{equation}
\begin{array}{l}
\quad \,\, \mathcal{L}_{\text{DPO}} = -\mathbb{E}_{(x,y_{l},y_{w}) \sim \mathcal{D}} \\[0.3em]
\left[ \log \sigma
\left(
\beta \log \frac{\pi(y_{w} | x)}{\pi_{\text{ref}}(y_{w} | x)}
- \beta \log \frac{\pi(y_{l} | x)}{\pi_{\text{ref}}(y_{l} | x)}
\right) \right] ,
\end{array}
\label{eq:dpo}
\end{equation}
where $\sigma(\ast)$ denotes the logistic function, $\beta=0.1$ is a hyperparameter, and $\pi_\text{ref}$ is the frozen reference model typically the model after SFT training. 

Despite these efforts, pilot experiments suggest that small LLMs struggle with generating good responses for complex (level-3) instructions.
To mitigate this issue, we adopt a progressive training strategy, scaling from simpler (level-1) to more complex (level-3) instructions.
This strategy can maximize the sampling efficiency in collecting self-generated data and ensure the consistent improvements, since each level's training is based on the trained checkpoint of the last level. 
After this progressive training, the small LLM is expected to follow instructions more precisely, making it better suited for LLM-based applications. The full pipeline of the proposed method is listed in Alg.~\ref{alg:enhance}.

\begin{table*}[t]
\centering
\small
\resizebox{0.97\linewidth}{!}{
\begin{tabular}{l|lll|ll|lll} 

\toprule

\multirow{2}{*}{Model} & \multicolumn{3}{c|}{\dataset}  & \multicolumn{2}{c|}{IFEval} & \multicolumn{3}{c}{InfoBench}                  \\
                       & level-1 & level-2 & level-3  & Prompt & Instruction & \multicolumn{1}{l}{Easy} & Hard  & Avg.  \\ 
\midrule
GPT-3.5                & 62.93   & 34.07   & 16.40   & 56.56        & 67.51        &  -                        &  -  & 86.71*       \\
GPT-4-turbo            & 76.29   & 53.33   & 35.31    & 79.71        & 85.67        &                        -  &    -   & 89.42*      \\
\midrule
LLaMA-2-13B            & 48.08   & 21.40   & 9.65     & 33.00        & 44.24        & 80.40                    & 77.10 & 78.12      \\
LLaMA-2-70B            & 55.57   & 26.47   & 11.89    & 44.36       &54.43               & 81.28                     &79.87     &80.30      \\
LLaMA-3-70B            & 65.63   & 40.02   & 23.55   &77.81      &84.30           & 86.07                     &86.80     &86.61          \\
Qwen-1.5-7B            & 58.66   & 29.49   & 13.87   & 39.00       & 50.96        & 77.82                         &75.13    &75.95      \\
WizardLM-7B            & 55.37   & 28.63   & 14.00    & 43.25        & 55.63      & 80.58                    & 77.70 & 78.58        \\
\midrule
\midrule

Mistral-7B             & 52.18   & 22.82   & 9.49     & \textbf{40.85}        & 50.84     & \textbf{76.67}                    & \textbf{71.15} & \textbf{72.84}         \\
Mistral-7B (ours)       & \textbf{61.85}   & \textbf{32.66}   & \textbf{15.97}      & 37.50        & \textbf{51.19} &72.17                      &68.25      &70.48              \\
\midrule
LLaMA-2-7B             & 50.91   & 22.01   & 9.31    & 31.42        & 44.96          & 18.43                    & 10.57 & 12.98     \\
LLaMA-2-7B (ours)        & \textbf{57.59}   &\textbf{27.72}   & \textbf{13.28}    & \textbf{40.48}        & \textbf{54.08}          & \textbf{73.00}                    & \textbf{68.08} & \textbf{69.59}    \\
\midrule
LLaMA-3-8B             & 60.36   & 31.86   & 15.81    & 68.22        & 77.14         & \textbf{81.88}                    & \textbf{83.72} & \textbf{83.10}     \\
LLaMA-3-8B (ours)       & \textbf{85.56}   & \textbf{59.67}   & \textbf{38.36}   &\textbf{68.50}        & \textbf{77.24}  & 79.10                    & 76.50 & 78.13             \\

\bottomrule
\end{tabular}
}

\caption{Results of various LLMs on three benchmarks. The best performance is highlighted in bold. Results with are from~\citet{sun2024conifer}. The strict mode is adopted for IFEval benchmark. IFEval and InfoBench are testing the performance of out-of-domain format following and general instruction following, respectively.}
\label{tab:main_table}
\end{table*}

\section{Experiment}
In this section, we take experiments to comprehensively validate the effectiveness of our method in enhancing 7B level LLMs' format control ability.

\subsection{Experimental Setup}
\paragraph{Data.}
For each level of \dataset dataset, we curate $10$k and $7$k samples for training and testing, respectively.
Considering the training costs and diversity, we do not fully extend this data. Here, the ideal maximum numbers of producible instructions for each level of \dataset  are about $60$k, $360$k and $2160$k.

\paragraph{Benchmarks.}
We validate our method on two instruction following benchmarks: (1) InfoBench~\citep{qin2024infobench}, which utilizes GPT-4-based evaluations to test the general instruction following ability. 
(2) IFEval~\citep{ifeval}, which adopts Python-based function evaluations for evaluating the format control ability, similar to our \dataset. 
Note that both benchmarks consist of approximately $500$ test samples, which may introduce higher variance and bias compared to our dataset.

\paragraph{Baselines.}
For comparison, we select various open-source LLMs, including Mistral~\citep{jiang2023mistral}, the LLaMA family of models~\citep{llama2, llama3}, Qwen~\citep{qwen}, and WizardLM~\citep{xu2023wizardlm}, all of which have been fine-tuned for alignment with human preferences. In addition, we also include GPT-3.5 and GPT-4 as reference baselines to show the gap between small LLMs and advanced LLMs.

\paragraph{Training Settings}
In this paper, our experiments follow the default settings of most hyperparameters in the SFTTrainer and DPOTrainer from LLaMA-Factory~\citep{zheng2024llamafactory}. The models are trained for a total of 8 epochs using a batch size of 4 on NVIDIA A6000 GPUs, which will take up to 1 hour. We employ the AdamW optimizer~\citep{loshchilov2018decoupled} with a learning rate of $5e-6$, coupled with a cosine learning rate scheduler. To accelerate the training and save the computation resources, we fine-tune the LLM with the LoRA~\citep{hu2021lora} adapter for which we set the rank and $\alpha$ to 64 and 128, respectively. Despite the query and value heads of attention blocks, all other parameters are frozen.

\subsection{Main Results}
The main results are shown in Table~\ref{tab:main_table}.

\paragraph{Prevalent Limitations.}
The results first suggest that 7B level open-source LLMs struggle with level-2 and level-3 format following instructions of our \dataset dataset and IFEval, while they commonly have acceptable performance on general instruction following dataset InfoBench. This clearly demonstrates the limitations of such small LLMs in adhering to specific formats.

\paragraph{Effectiveness of Enhancing Format Control.}
Due to limited resources, we select only three popular LLMs for training. Their superior performance on both in-domain data (\dataset) and out-of-domain data (IFEval) confirms the effectiveness of the proposed method in enhancing format following ability. 
Notably, the trained LLaMA-3-8B model outperforms GPT-4 in the level-3 format following task.
However, we observed a slight decline for Mistral and LLaMA-3-8B in the general instruction following data (InfoBench) though LLaMA-2-7B greatly improves.
This may be due to:
(1) Overfitting on general instruction following tasks (they have significantly better performance than LLaMA-2-7B), where training on format following may slightly affect performance. 
(2) The limited number of test samples and LLM-based evaluations, both of which introduce additional evaluation bias.

\section{Analysis}

\begin{table}
\small
\centering
\begin{tabular}{l|ll|ll|ll} 
\toprule
 Metric & \multicolumn{2}{c|}{level-1} & \multicolumn{2}{c|}{level-2} & \multicolumn{2}{c}{level-3}  \\
    & G.  & H.                    & G.  & H.                    & G.  & H.                     \\ 
\midrule
Conflict $\downarrow$  & 1.5 & 1.0                   & 1.9 & 1.5                   & 2.5 & 2.0                    \\
Reasonable $\uparrow$  & 3.6 & 4.3                   & 3.4 & 3.5                   & 2.9 & 3.3                    \\
Difficulty  & 2.3 & 1.8                   & 3.0 & 3.2                   & 3.5 & 4.0                    \\
\bottomrule
\end{tabular}
\caption{Human (H.) and GPT-4 (G.) evaluations (5 points) on the data quality of our \dataset dataset.}
\label{tab:human-gpt}
\end{table}

\subsection{Quality of \dataset dataset}
The instruction of \dataset is made by randomly pairing  the question and constraints, which may introduce the conflicts in the content. We sample $100$ instructions for manual and GPT-4 evaluations shown in Table~\ref{tab:human-gpt}. The results suggest that the curated instruction remain reasonable and have low conflicts. The difficulty of the instruction is consistent with the number of added constraints.

\begin{table}[t]
    \small
    \centering
    \begin{tabular}{l|ccc}
    \toprule
        Sampling & Accuracy & \# Correct Data & \# DPO Pair  \\ 
        \midrule
        \nth{1}  & 70.40 & 6945 / 9865  & 830  \\ 
        \nth{2} & 23.05 & 673 / 2920 & 157  \\ 
        \nth{3}  & 4.63 & 104 / 2247 & 53  \\ 
        \nth{4} & 2.47 & 53 / 2143 & -  \\ 
        Total & 78.80 & 7775 / 9865 & 1040  \\
        \midrule
        Direct &55.14    & 5440 / 9865 & 138 \\  
    \bottomrule
    \end{tabular}
    \caption{Comparison of different sampling strategies of level-1 training data collection. ``Direct'' means sampling multiple times without demonstration.}
     \label{tab:ab-sampling}
\end{table}

\subsection{Analysis of Enhancing Format Control}

\paragraph{Sampling Efficiency.} 
Table~\ref{tab:ab-sampling} shows details of the adopted one-shot demonstration in response sampling, which indicate that the self-generated wrong example can serve as a useful demonstration to help LLMs to generate better responses, leading to higher sampling efficiency than directly sampling multiple times.
\begin{table}[t]
    \small
    \centering
    \begin{tabular}{l|ccc}
    \toprule
    \textbf{Method} & \textbf{level-1} & \textbf{level-2} & \textbf{level-3} \\ \midrule
    Llama-3-8B & 59.75 & 32.19 & 16.66 \\ \midrule
    L1\textsubscript{SFT-Only} & 61.22 & 30.94 & 15.20 \\ 
    L1\textsubscript{DPO-Only} & 63.26 & 37.14 & 20.29 \\ 
    L1\textsubscript{SFT-DPO} & 77.96 & 50.68 & 30.00 \\ \midrule
    L2\textsubscript{SFT-Only} & 76.58 & 50.87 & 29.43 \\ 
    L2\textsubscript{DPO-Only} & 62.71 & 34.81 & 18.80 \\ 
    L2\textsubscript{SFT-DPO} & 82.79 & 56.94 & 35.44 \\ \midrule
    L3\textsubscript{SFT-Only} & 79.86 & 53.56 & 32.26 \\ 
    L3\textsubscript{DPO-Only} & 62.25 & 34.73 & 17.95 \\ 
    L3\textsubscript{SFT-DPO} & \textbf{85.56} & \textbf{59.67} & \textbf{38.36} \\ \bottomrule
    \end{tabular}
    \caption{Comparison of different training strategies in the progressive training procedure.}
    \label{tab:ab-progressive}
\end{table}

\paragraph{Training Strategies.}
As shown in Table~\ref{tab:ab-progressive}, we compare different training strategies with the progressive training strategy. First, the SFT training can effectively improve the format control performance, while the gains for level-2 and level-3 training are decreasing. For DPO-Only training, it will even harm the performance as the training proceeds. For the adopted strategy (SFT-DPO), the progressive training can consistently enhance the LLM's format following performance.

\subsection{LLM-based v.s. Python-based Judgment}

\begin{table}[t]
    \small
    \centering
    \begin{tabular}{l|lll}
    \toprule
        Method & Accuracy  &  Time(s) & Cost(\$)  \\
        \midrule
        GPT-4o-mini & 59.0  & 99.53  & 0.0144\\ 
        GPT-4o & 70.0 & 205.10  & 2.3830 \\ 
        \midrule
        Python  & \textbf{100.0 }  & \textbf{0.52} & \textbf{0.0000} \\ 
        
    \bottomrule
    \end{tabular}
     \caption{Human comparison of LLMs-based against our Python-based evaluations in terms of Accuracy, Time, and Cost on 200 samples.}
     \label{tab:gpt4omini}
\end{table}

\begin{figure*}[t]
\centering
\includegraphics[width=\linewidth]{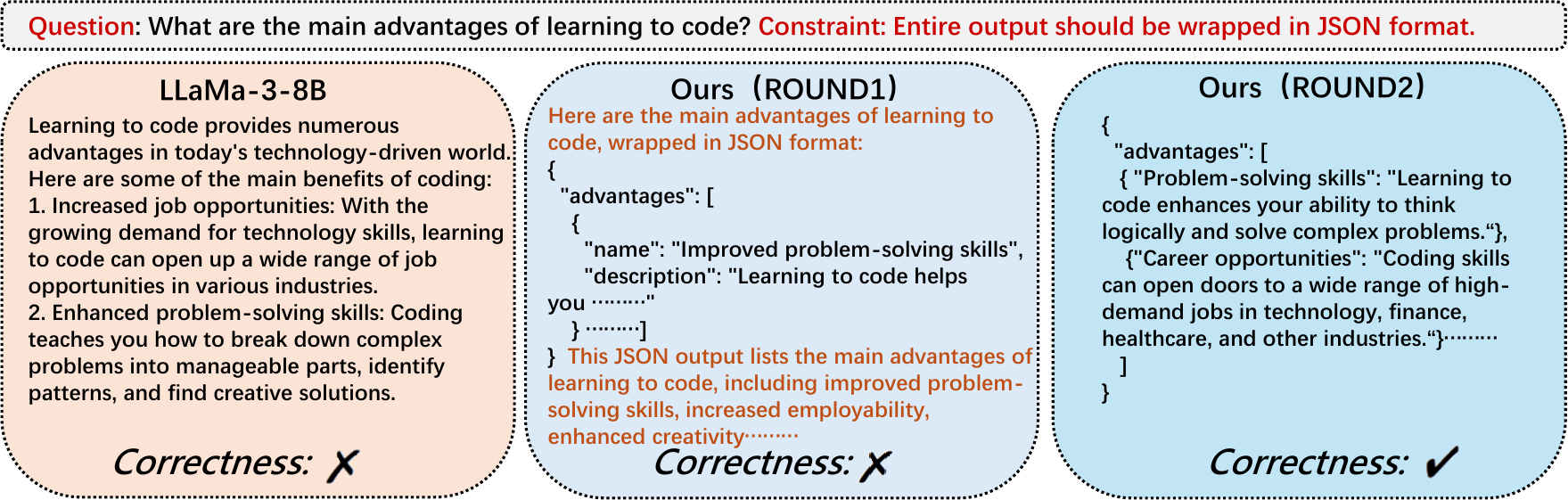}
\caption{Example of the generated responses to the question with JSON format constraint. }
\label{fig:json_case}
\end{figure*}

\paragraph{Accuracy \& Time \& Cost.}
Recent advancements in lightweight LLMs such as GPT-4o-mini motivate the community to choose LLM-based evaluations~\citep{qin2024infobench,jiang2023followbench,chan2023chateval}.
We compare the LLM-based judgment with Python-based judgment in Table~\ref{tab:gpt4omini}.
The results show that in the format following evaluation scenario, Python based method has a significant advantage over LLM-based approach even with lightweight LLMs (GPT-4o-mini), especially considering the 100x speed up.

\begin{table}[t]
    \small
    \centering
    \resizebox{1.0\columnwidth}{!}{
    \begin{tabular}{l|l|cccc}
    \toprule
        \multirow{2}{*}{Model} & \multirow{2}{*}{Metrics} & \multicolumn{4}{c}{Sampling Temperature}  \\
              &   & 0.1  & 0.3  & 0.7  & 1.0\\ 
        \midrule
        \multirow{3}{*}{GPT-4o} & Inc & 25\%   & 24\% &  33\%  & 48\%\\ 
          & Flip & 10.06   & 7.75  &  10.15  & 7.58\\ 
         \midrule
        \multirow{3}{*}{GPT-4o-mini} & Inc & 33\%   & 36\%  &  44\%  & 52\% \\ 
          & Flip & 11.06   & 9.14  &  8.75  & 12.34\\ 

    \bottomrule
    \end{tabular}
    }
         \caption{Consistency measurement. The model is queried 50 times for the same question and response using different temperatures. ``Inc'' and ``Flip'' denote average inconsistency rate across the set and count of judgment flips across the queries, respectively.
         }
    
     \label{tab:consistency}
\end{table}

\paragraph{Consistency of LLM-based Judgment.} 
We further show the inconsistency and instability of using LLM-based judgments for format following in Table~\ref{tab:consistency}. The results suggest that LLMs are not reliable even by setting the temperature to $0.1$. Moreover, the advanced GPT-4o is still inconsistent in judging the correctness of format following, which shows the limitations of LLM-based evaluations.

\begin{table}[t!]
    \small
    \centering

    \resizebox{1.0\columnwidth}{!}{
    \begin{tabular}{l|l|ccc|c}
    \toprule
        Model & Method & level-1  &  level-2 & level-3 & IFEval \\
        \midrule
        \multirow{2}{*}{LLaMa-3-8B} & Human & 2.52  &  1.87 & 1.58  &2.43\\
                                    & GPT-4o & 2.73  &  1.37 & 1.08  & 2.17\\
        \midrule
        \multirow{2}{*}{Ours} & Human & 3.69  &  2.81 & 2.10  & 3.58\\
                                    & GPT-4o & 4.32  &  3.58 & 2.27  & 4.15\\
    \bottomrule
    \end{tabular}
    }

\caption{Quality assessment for 200 generated responses in each dataset. The score spreads out on a scale of 0-5.}
\label{tab:content_score}

\end{table}

\subsection{Quality of Generated Responses}
We use manual and GPT-4o evaluations to assess the response quality considering both the content and format in Table~\ref{tab:content_score}. The first evaluation criteria is whether the generated content is relevant to the input and whether the requirements of the input are fulfilled with high quality. The second criterion is whether the generated content satisfies the constraints added in the input. 
From the results, we surprisingly observed that the LLM trained with our method not only improves the completion of constraints, but also enhances the quality of the content generation.
Additionally, we find that our method can achieve a satisfactory performance on out-of-domain IFEval data, indicating its superior generalizatio in format following.

\paragraph{Case Study.} To intuitively understand the effects of the proposed progressive training, Figure~\ref{fig:json_case} presents an example of different methods responding to JSON format instruction.
First, we find that the LLaMA-3-8B fails to follow the desired JSON output format.
However, the model trained with our level-1 constraints (ROUND1) shows noticeable improvements in generating outputs with better format control, though it still does not fully adhere to the given instructions.
Further training on level-2 data (ROUND2) can also help improve its format following ability, demonstrating the effectiveness of our progressive training method.

\section{Conclusion}
In this paper, we focus on enhancing the format following ability of 7B level LLMs.
First, to evaluate the format following ability of LLMs, we curate a fully verifiable format following dataset \dataset which uses Python scripts to accurately judge the correctness of the format following.
Then, by leveraging the verifiable feature, we can synthesize massive training data to enhance such format control generations of small LLMs in a self-improvement paradigm.
Our experiments and analysis reveal the limitations of small LLMs in format following, while demonstrating the effectiveness of our method in improving format control generations.
We believe these findings will benefit the  research community and advance the LLM applications.

\clearpage
\newpage
\section*{Limitations}
In this section, we discuss the observed limitations and offer useful suggestions for future research.
\begin{enumerate}[1)]
    \item The contributions of this paper heavily rely on the fact that output formats can be efficiently verified by Python functions, thus it may not be easily extended to general instruction following. Fine-grained format control remains a significant challenge for current open-source 7B-level LLMs. Moreover, generating specific output formats is crucial for many LLM-based applications. Although constrained decoding is promising, it can be further improved by supporting a broader range of formats.
    \item The richness of the proposed verifiable format following dataset \dataset is based on about $60$ meta constraints which may not cover the whole categories of human desired output formats in real world applications. In the future, it can be integrated with online human feedback to collect more format categories. 
    \item The adopted training methods for improving LLM's format control ability mainly use the supervised fine-tuning and DPO training~\citep{rafailov2024direct}, which leverage the verifiable feature of our \dataset dataset for annotations. However, the verifiable rule can be viewed as a reward function to explore reinforcement learning algorithms for further improvements.
    \item Due to the limited computation resources, the proposed training method is validated on only three 7B-level LLMs (i.e., Mistral~\citep{jiang2023mistral}, LLaMA-2~\citep{llama2} and LLaMA-3~\citep{llama3}). Also, we only evaluate on other two instruction following benchmarks (i.e., InfoBench~\citep{qin2024infobench} and IFEval~\citep{ifeval}). And the analysis can be more comprehensive by exploring the relationship between training data size and the performance.  Besides, the results indicate that our training method may slightly harm some general instruction following performance, which needs more investigation.
\end{enumerate}


\section*{Ethics Statement}
All datasets and trained LLMs employed in this paper are publicly available. This paper mainly studies the format following issue of LLMs, while does not cover issues of 
evaluating the correctness of the content such as detecting hallucinations. This indicates that the proposed training method may not enhance other foundational abilities of LLMs.
We use ChatGPT at the sentence level (e.g., fixing grammar) to assist the paper writing.


\bibliography{custom,anthology}

\begin{thebibliography}{53}
\providecommand{\natexlab}[1]{#1}

\bibitem[{{1rgs}(2023)}]{jsonformer2023}
{1rgs}. 2023.
\newblock \href {https://github.com/1rgs/jsonformer} {jsonformer - github repository}.

\bibitem[{Bai et~al.(2023)Bai, Bai, Chu, Cui, Dang, Deng, Fan, Ge, Han, Huang, Hui, Ji, Li, Lin, Lin, Liu, Liu, Lu, Lu, Ma, Men, Ren, Ren, Tan, Tan et~al.}]{qwen}
Jinze Bai, Shuai Bai, Yunfei Chu, Zeyu Cui, Kai Dang, Xiaodong Deng, Yang Fan, Wenbin Ge, Yu~Han, Fei Huang, Binyuan Hui, Luo Ji, Mei Li, Junyang Lin, Runji Lin, Dayiheng Liu, Gao Liu, Chengqiang Lu, Keming Lu, Jianxin Ma, Rui Men, Xingzhang Ren, Xuancheng Ren, Chuanqi Tan, Sinan Tan, et~al. 2023.
\newblock Qwen technical report.
\newblock \emph{arXiv preprint arXiv:2309.16609}.

\bibitem[{Bai et~al.(2022)Bai, Jones, Ndousse, Askell, Chen, DasSarma, Drain, Fort, Ganguli, Henighan, Joseph, Kadavath, Kernion, Conerly, El-Showk, Elhage, Hatfield-Dodds, Hernandez, Hume, Johnston, Kravec, Lovitt, Nanda, Olsson, Amodei, Brown, Clark, McCandlish, Olah, Mann, and Kaplan}]{bai2022training}
Yuntao Bai, Andy Jones, Kamal Ndousse, Amanda Askell, Anna Chen, Nova DasSarma, Dawn Drain, Stanislav Fort, Deep Ganguli, Tom Henighan, Nicholas Joseph, Saurav Kadavath, Jackson Kernion, Tom Conerly, Sheer El-Showk, Nelson Elhage, Zac Hatfield-Dodds, Danny Hernandez, Tristan Hume, Scott Johnston, Shauna Kravec, Liane Lovitt, Neel Nanda, Catherine Olsson, Dario Amodei, Tom Brown, Jack Clark, Sam McCandlish, Chris Olah, Ben Mann, and Jared Kaplan. 2022.
\newblock Training a {Helpful} and {Harmless} {Assistant} with {Reinforcement} {Learning} from {Human} {Feedback}.
\newblock \emph{arXiv}, abs/2204.05862.

\bibitem[{Bang et~al.(2023)Bang, Cahyawijaya, Lee, Dai, Su, Wilie, Lovenia, Ji, Yu, Chung et~al.}]{bang2023multitask}
Yejin Bang, Samuel Cahyawijaya, Nayeon Lee, Wenliang Dai, Dan Su, Bryan Wilie, Holy Lovenia, Ziwei Ji, Tiezheng Yu, Willy Chung, et~al. 2023.
\newblock A multitask, multilingual, multimodal evaluation of chatgpt on reasoning, hallucination, and interactivity.
\newblock \emph{arXiv preprint arXiv:2302.04023}.

\bibitem[{Brown et~al.(2020)Brown, Mann, Ryder, Subbiah, Kaplan, Dhariwal, Neelakantan, Shyam, Sastry, Askell, Agarwal, Herbert-Voss, Krueger, Henighan, Child, Ramesh, Ziegler, Wu, Winter, Hesse, Chen, Sigler, Litwin, Gray, Chess, Clark, Berner, McCandlish, Radford, Sutskever, and Amodei}]{brown2020language}
Tom~B. Brown, Benjamin Mann, Nick Ryder, Melanie Subbiah, Jared Kaplan, Prafulla Dhariwal, Arvind Neelakantan, Pranav Shyam, Girish Sastry, Amanda Askell, Sandhini Agarwal, Ariel Herbert-Voss, Gretchen Krueger, Tom Henighan, Rewon Child, Aditya Ramesh, Daniel~M. Ziegler, Jeffrey Wu, Clemens Winter, Christopher Hesse, Mark Chen, Eric Sigler, Mateusz Litwin, Scott Gray, Benjamin Chess, Jack Clark, Christopher Berner, Sam McCandlish, Alec Radford, Ilya Sutskever, and Dario Amodei. 2020.
\newblock Language {Models} are {Few}-{Shot} {Learners}.
\newblock In \emph{Conference on {Neural} {Information} {Processing} {Systems} ({NeurIPS})}.

\bibitem[{Bubeck et~al.(2023)Bubeck, Chandrasekaran, Eldan, Gehrke, Horvitz, Kamar, Lee, Lee, Li, Lundberg et~al.}]{bubeck2023sparks}
S{\'e}bastien Bubeck, Varun Chandrasekaran, Ronen Eldan, Johannes Gehrke, Eric Horvitz, Ece Kamar, Peter Lee, Yin~Tat Lee, Yuanzhi Li, Scott Lundberg, et~al. 2023.
\newblock Sparks of artificial general intelligence: Early experiments with gpt-4.
\newblock \emph{arXiv preprint arXiv:2303.12712}.

\bibitem[{Chan et~al.(2023)Chan, Chen, Su, Yu, Xue, Zhang, Fu, and Liu}]{chan2023chateval}
Chi-Min Chan, Weize Chen, Yusheng Su, Jianxuan Yu, Wei Xue, Shanghang Zhang, Jie Fu, and Zhiyuan Liu. 2023.
\newblock Chateval: Towards better llm-based evaluators through multi-agent debate.
\newblock \emph{arXiv preprint arXiv:2308.07201}.

\bibitem[{Chiang and Lee(2023)}]{chiang-lee-2023-large}
Cheng-Han Chiang and Hung-yi Lee. 2023.
\newblock \href {https://doi.org/10.18653/v1/2023.acl-long.870} {Can large language models be an alternative to human evaluations?}
\newblock In \emph{Proceedings of the 61st Annual Meeting of the Association for Computational Linguistics (Volume 1: Long Papers)}, pages 15607--15631, Toronto, Canada. Association for Computational Linguistics.

\bibitem[{Chiang et~al.(2023)Chiang, Li, Lin, Sheng, Wu, Zhang, Zheng, Zhuang, Zhuang, Gonzalez, Stoica, and Xing}]{vicuna2023}
Wei-Lin Chiang, Zhuohan Li, Zi~Lin, Ying Sheng, Zhanghao Wu, Hao Zhang, Lianmin Zheng, Siyuan Zhuang, Yonghao Zhuang, Joseph~E. Gonzalez, Ion Stoica, and Eric~P. Xing. 2023.
\newblock \href {https://lmsys.org/blog/2023-03-30-vicuna/} {Vicuna: An open-source chatbot impressing gpt-4 with 90\%* chatgpt quality}.

\bibitem[{Dong et~al.(2023)Dong, Xiong, Goyal, Pan, Diao, Zhang, Shum, and Zhang}]{dong2023raft}
Hanze Dong, Wei Xiong, Deepanshu Goyal, Rui Pan, Shizhe Diao, Jipeng Zhang, Kashun Shum, and T.~Zhang. 2023.
\newblock Raft: Reward {rAnked} {FineTuning} for {Generative} {Foundation} {Model} {Alignment}.
\newblock \emph{Transactions on Machine Learning Research}, abs/2304.06767.

\bibitem[{Dong et~al.(2024)Dong, Ruan, Cai, Lai, Xu, Zhao, and Chen}]{dong2024xgrammar}
Yixin Dong, Charlie~F Ruan, Yaxing Cai, Ruihang Lai, Ziyi Xu, Yilong Zhao, and Tianqi Chen. 2024.
\newblock Xgrammar: Flexible and efficient structured generation engine for large language models.
\newblock \emph{arXiv preprint arXiv:2411.15100}.

\bibitem[{{ETH SRI}(2023)}]{lmql2023}
{ETH SRI}. 2023.
\newblock \href {https://github.com/eth-sri/lmql} {Lmql - github repository}.

\bibitem[{Ethayarajh et~al.(2024)Ethayarajh, Xu, Muennighoff, Jurafsky, and Kiela}]{ethayarajh2024kto}
Kawin Ethayarajh, Winnie Xu, Niklas Muennighoff, Dan Jurafsky, and Douwe Kiela. 2024.
\newblock Kto: Model alignment as prospect theoretic optimization.
\newblock \emph{arXiv preprint arXiv:2402.01306}.

\bibitem[{Fu et~al.(2023)Fu, Ng, Jiang, and Liu}]{fu2023gptscore}
Jinlan Fu, See-Kiong Ng, Zhengbao Jiang, and Pengfei Liu. 2023.
\newblock Gptscore: Evaluate as you desire.
\newblock \emph{arXiv}, abs/2302.04166.

\bibitem[{Gou et~al.(2024)Gou, Shao, Gong, yelong shen, Yang, Duan, and Chen}]{gou2024critic}
Zhibin Gou, Zhihong Shao, Yeyun Gong, yelong shen, Yujiu Yang, Nan Duan, and Weizhu Chen. 2024.
\newblock \href {https://openreview.net/forum?id=Sx038qxjek} {{CRITIC}: Large language models can self-correct with tool-interactive critiquing}.
\newblock In \emph{The Twelfth International Conference on Learning Representations}.

\bibitem[{He et~al.(2024)He, Zeng, He, Liang, and Xiao}]{he2024complex}
Qianyu He, Jie Zeng, Qianxi He, Jiaqing Liang, and Yanghua Xiao. 2024.
\newblock From complex to simple: Enhancing multi-constraint complex instruction following ability of large language models.
\newblock \emph{arXiv preprint arXiv:2404.15846}.

\bibitem[{Hong et~al.(2024)Hong, Lee, and Thorne}]{hong2024orpo}
Jiwoo Hong, Noah Lee, and James Thorne. 2024.
\newblock Orpo: Monolithic preference optimization without reference model.
\newblock \emph{arXiv preprint arXiv:2403.07691}, 2(4):5.

\bibitem[{Hu et~al.(2021)Hu, Shen, Wallis, Allen-Zhu, Li, Wang, Wang, and Chen}]{hu2021lora}
Edward~J Hu, Yelong Shen, Phillip Wallis, Zeyuan Allen-Zhu, Yuanzhi Li, Shean Wang, Lu~Wang, and Weizhu Chen. 2021.
\newblock Lora: Low-rank adaptation of large language models.
\newblock \emph{arXiv preprint arXiv:2106.09685}.

\bibitem[{Huang et~al.(2023)Huang, Gu, Hou, Wu, Wang, Yu, and Han}]{huang-etal-2023-large}
Jiaxin Huang, Shixiang Gu, Le~Hou, Yuexin Wu, Xuezhi Wang, Hongkun Yu, and Jiawei Han. 2023.
\newblock \href {https://doi.org/10.18653/v1/2023.emnlp-main.67} {Large language models can self-improve}.
\newblock In \emph{Proceedings of the 2023 Conference on Empirical Methods in Natural Language Processing}, pages 1051--1068, Singapore. Association for Computational Linguistics.

\bibitem[{Jain et~al.(2023)Jain, Saifullah, Wen, Kirchenbauer, Shu, Saha, Goldblum, Geiping, and Goldstein}]{jain2023bring}
Neel Jain, Khalid Saifullah, Yuxin Wen, John Kirchenbauer, Manli Shu, Aniruddha Saha, Micah Goldblum, Jonas Geiping, and Tom Goldstein. 2023.
\newblock Bring your own data! self-supervised evaluation for large language models.
\newblock \emph{arXiv preprint arXiv:2306.13651}.

\bibitem[{Jiang et~al.(2023{\natexlab{a}})Jiang, Sablayrolles, Mensch, Bamford, Chaplot, Casas, Bressand, Lengyel, Lample, Saulnier et~al.}]{jiang2023mistral}
Albert~Q Jiang, Alexandre Sablayrolles, Arthur Mensch, Chris Bamford, Devendra~Singh Chaplot, Diego de~las Casas, Florian Bressand, Gianna Lengyel, Guillaume Lample, Lucile Saulnier, et~al. 2023{\natexlab{a}}.
\newblock Mistral 7b.
\newblock \emph{arXiv preprint arXiv:2310.06825}.

\bibitem[{Jiang et~al.(2023{\natexlab{b}})Jiang, Wang, Zeng, Zhong, Li, Mi, Shang, Jiang, Liu, and Wang}]{jiang2023followbench}
Yuxin Jiang, Yufei Wang, Xingshan Zeng, Wanjun Zhong, Liangyou Li, Fei Mi, Lifeng Shang, Xin Jiang, Qun Liu, and Wei Wang. 2023{\natexlab{b}}.
\newblock Followbench: A multi-level fine-grained constraints following benchmark for large language models.
\newblock \emph{arXiv preprint arXiv:2310.20410}.

\bibitem[{Jiang et~al.(2023{\natexlab{c}})Jiang, Huang, and Wei}]{jiang2023preference}
Zaifan Jiang, Xing Huang, and Chao Wei. 2023{\natexlab{c}}.
\newblock Preference as reward, maximum preference optimization with importance sampling.
\newblock \emph{arXiv preprint arXiv:2312.16430}.

\bibitem[{Lin and Chen(2023)}]{lin-chen-2023-llm}
Yen-Ting Lin and Yun-Nung Chen. 2023.
\newblock \href {https://doi.org/10.18653/v1/2023.nlp4convai-1.5} {{LLM}-eval: Unified multi-dimensional automatic evaluation for open-domain conversations with large language models}.
\newblock In \emph{Proceedings of the 5th Workshop on NLP for Conversational AI (NLP4ConvAI 2023)}, pages 47--58, Toronto, Canada. Association for Computational Linguistics.

\bibitem[{Liu et~al.(2023{\natexlab{a}})Liu, Iter, Xu, Wang, Xu, and Zhu}]{liu2303g}
Yang Liu, Dan Iter, Yichong Xu, Shuohang Wang, Ruochen Xu, and Chenguang Zhu. 2023{\natexlab{a}}.
\newblock G-eval: Nlg evaluation using gpt-4 with better human alignment (2023).
\newblock \emph{URL http://arxiv. org/abs/2303.16634}.

\bibitem[{Liu et~al.(2023{\natexlab{b}})Liu, Yang, Huang, Zhang, Huang, Wei, Deng, Sun, and Zhang}]{liu2023calibrating}
Yuxuan Liu, Tianchi Yang, Shaohan Huang, Zihan Zhang, Haizhen Huang, Furu Wei, Weiwei Deng, Feng Sun, and Qi~Zhang. 2023{\natexlab{b}}.
\newblock Calibrating llm-based evaluator.
\newblock \emph{arXiv}, abs/2309.13308.

\bibitem[{Longpre et~al.(2023)Longpre, Hou, Vu, Webson, Chung, Tay, Zhou, Le, Zoph, Wei et~al.}]{longpre2023flan}
Shayne Longpre, Le~Hou, Tu~Vu, Albert Webson, Hyung~Won Chung, Yi~Tay, Denny Zhou, Quoc~V Le, Barret Zoph, Jason Wei, et~al. 2023.
\newblock The flan collection: Designing data and methods for effective instruction tuning.
\newblock In \emph{International Conference on Machine Learning}, pages 22631--22648. PMLR.

\bibitem[{Loshchilov and Hutter(2019)}]{loshchilov2018decoupled}
Ilya Loshchilov and Frank Hutter. 2019.
\newblock \href {https://openreview.net/forum?id=Bkg6RiCqY7} {Decoupled weight decay regularization}.
\newblock In \emph{International Conference on Learning Representations}.

\bibitem[{Ma et~al.(2024)Ma, Zhan, Wong, and Chao}]{ma2024activate}
Jingkun Ma, Runzhe Zhan, Derek~F Wong, and Lidia~S Chao. 2024.
\newblock Activate integrated controllable generation with soft prompt.
\newblock In \emph{CCF International Conference on Natural Language Processing and Chinese Computing}, pages 239--251. Springer.

\bibitem[{Meng et~al.(2024)Meng, Xia, and Chen}]{meng2024simpo}
Yu~Meng, Mengzhou Xia, and Danqi Chen. 2024.
\newblock Simpo: Simple preference optimization with a reference-free reward.
\newblock \emph{arXiv preprint arXiv:2405.14734}.

\bibitem[{Mishra et~al.(2021)Mishra, Khashabi, Baral, and Hajishirzi}]{mishra2021cross}
Swaroop Mishra, Daniel Khashabi, Chitta Baral, and Hannaneh Hajishirzi. 2021.
\newblock Cross-task generalization via natural language crowdsourcing instructions.
\newblock \emph{arXiv preprint arXiv:2104.08773}.

\bibitem[{OpenAI(2023)}]{openai2023gpt4}
OpenAI. 2023.
\newblock \href {https://arxiv.org/abs/2303.08774} {Gpt-4 technical report}.
\newblock \emph{arXiv}, abs/2303.08774.

\bibitem[{{OpenAI}(2023)}]{openai2023structured}
{OpenAI}. 2023.
\newblock \href {https://openai.com/index/introducing-structured-outputs-in-the-api} {Introducing structured outputs in the api}.

\bibitem[{{Outlines Development Team}(2023)}]{outlines2023}
{Outlines Development Team}. 2023.
\newblock \href {https://github.com/outlines-dev/outlines} {Outlines - github repository}.

\bibitem[{Ouyang et~al.(2022)Ouyang, Wu, Jiang, Almeida, Wainwright, Mishkin, Zhang, Agarwal, Slama, Ray et~al.}]{ouyang2022training}
Long Ouyang, Jeffrey Wu, Xu~Jiang, Diogo Almeida, Carroll Wainwright, Pamela Mishkin, Chong Zhang, Sandhini Agarwal, Katarina Slama, Alex Ray, et~al. 2022.
\newblock Training language models to follow instructions with human feedback.
\newblock \emph{Advances in neural information processing systems}, 35:27730--27744.

\bibitem[{Qin et~al.(2024)Qin, Song, Hu, Yao, Cho, Wang, Wu, Liu, Liu, and Yu}]{qin2024infobench}
Yiwei Qin, Kaiqiang Song, Yebowen Hu, Wenlin Yao, Sangwoo Cho, Xiaoyang Wang, Xuansheng Wu, Fei Liu, Pengfei Liu, and Dong Yu. 2024.
\newblock \href {https://arxiv.org/abs/2401.03601} {Infobench: Evaluating instruction following ability in large language models}.
\newblock \emph{arXiv}.

\bibitem[{Rafailov et~al.(2024)Rafailov, Sharma, Mitchell, Manning, Ermon, and Finn}]{rafailov2024direct}
Rafael Rafailov, Archit Sharma, Eric Mitchell, Christopher~D Manning, Stefano Ermon, and Chelsea Finn. 2024.
\newblock Direct preference optimization: Your language model is secretly a reward model.
\newblock \emph{Advances in Neural Information Processing Systems}, 36.

\bibitem[{Sun et~al.(2024)Sun, Liu, Li, Wang, Dong, Lin, and Huang}]{sun2024conifer}
Haoran Sun, Lixin Liu, Junjie Li, Fengyu Wang, Baohua Dong, Ran Lin, and Ruohui Huang. 2024.
\newblock \href {https://arxiv.org/abs/2404.02823} {Conifer: Improving complex constrained instruction-following ability of large language models}.
\newblock \emph{arxiv preprint arXiv:2404.02823}.

\bibitem[{Taori et~al.(2023)Taori, Gulrajani, Zhang, Dubois, Li, Guestrin, Liang, and Hashimoto}]{alpaca}
Rohan Taori, Ishaan Gulrajani, Tianyi Zhang, Yann Dubois, Xuechen Li, Carlos Guestrin, Percy Liang, and Tatsunori~B. Hashimoto. 2023.
\newblock Stanford alpaca: An instruction-following llama model.
\newblock \url{https://github.com/tatsu-lab/stanford_alpaca}.

\bibitem[{Touvron et~al.(2023{\natexlab{a}})Touvron, Martin, Stone, Albert, Almahairi, Babaei, Bashlykov, Batra, Bhargava, Bhosale, Bikel, Blecher, Ferrer et~al.}]{llama2}
Hugo Touvron, Louis Martin, Kevin Stone, Peter Albert, Amjad Almahairi, Yasmine Babaei, Nikolay Bashlykov, Soumya Batra, Prajjwal Bhargava, Shruti Bhosale, Dan Bikel, Lukas Blecher, Cristian~Canton Ferrer, et~al. 2023{\natexlab{a}}.
\newblock Llama 2: Open foundation and fine-tuned chat models.
\newblock \emph{arXiv}, abs/2307.09288.

\bibitem[{Touvron et~al.(2023{\natexlab{b}})Touvron, Martin, Stone, Albert, Almahairi, Babaei, Bashlykov, Batra, Bhargava, Bhosale et~al.}]{touvron2023llama}
Hugo Touvron, Louis Martin, Kevin Stone, Peter Albert, Amjad Almahairi, Yasmine Babaei, Nikolay Bashlykov, Soumya Batra, Prajjwal Bhargava, Shruti Bhosale, et~al. 2023{\natexlab{b}}.
\newblock Llama 2: Open foundation and fine-tuned chat models.
\newblock \emph{arXiv preprint arXiv:2307.09288}.

\bibitem[{Touvron et~al.(2024)Touvron, Team et~al.}]{llama3}
Hugo Touvron, Meta~AI Team, et~al. 2024.
\newblock \href {https://ai.meta.com/blog/meta-llama-3/} {{Introducing Meta Llama 3: The most capable openly available LLM to date}}.

\bibitem[{Wang et~al.(2023{\natexlab{a}})Wang, Yu, Zeng, Yang, Wang, Chen, Jiang, Xie, Wang, Xie et~al.}]{wang2023pandalm}
Yidong Wang, Zhuohao Yu, Zhengran Zeng, Linyi Yang, Cunxiang Wang, Hao Chen, Chaoya Jiang, Rui Xie, Jindong Wang, Xing Xie, et~al. 2023{\natexlab{a}}.
\newblock Pandalm: An automatic evaluation benchmark for llm instruction tuning optimization.
\newblock \emph{arXiv preprint arXiv:2306.05087}.

\bibitem[{Wang et~al.(2023{\natexlab{b}})Wang, Kordi, Mishra, Liu, Smith, Khashabi, and Hajishirzi}]{wang-etal-2023-self-instruct}
Yizhong Wang, Yeganeh Kordi, Swaroop Mishra, Alisa Liu, Noah~A. Smith, Daniel Khashabi, and Hannaneh Hajishirzi. 2023{\natexlab{b}}.
\newblock \href {https://doi.org/10.18653/v1/2023.acl-long.754} {Self-instruct: Aligning language models with self-generated instructions}.
\newblock In \emph{Proceedings of the 61st Annual Meeting of the Association for Computational Linguistics (Volume 1: Long Papers)}, pages 13484--13508, Toronto, Canada. Association for Computational Linguistics.

\bibitem[{Wang et~al.(2023{\natexlab{c}})Wang, Huang, Liu, Wang, Song, Zhang, Huang, Wei, Deng, Sun, and Zhang}]{wang-etal-2023-democratizing}
Zhaoyang Wang, Shaohan Huang, Yuxuan Liu, Jiahai Wang, Minghui Song, Zihan Zhang, Haizhen Huang, Furu Wei, Weiwei Deng, Feng Sun, and Qi~Zhang. 2023{\natexlab{c}}.
\newblock \href {https://doi.org/10.18653/v1/2023.emnlp-main.120} {Democratizing reasoning ability: Tailored learning from large language model}.
\newblock In \emph{Proceedings of the 2023 Conference on Empirical Methods in Natural Language Processing}, pages 1948--1966, Singapore. Association for Computational Linguistics.

\bibitem[{Wei et~al.(2022)Wei, Tay, Bommasani, Raffel, Zoph, Borgeaud, Yogatama, Bosma, Zhou, Metzler, Chi, Hashimoto, Vinyals, Liang, Dean, and Fedus}]{wei2022emergentabilitieslargelanguage}
Jason Wei, Yi~Tay, Rishi Bommasani, Colin Raffel, Barret Zoph, Sebastian Borgeaud, Dani Yogatama, Maarten Bosma, Denny Zhou, Donald Metzler, Ed~H. Chi, Tatsunori Hashimoto, Oriol Vinyals, Percy Liang, Jeff Dean, and William Fedus. 2022.
\newblock \href {https://arxiv.org/abs/2206.07682} {Emergent abilities of large language models}.
\newblock \emph{Preprint}, arXiv:2206.07682.

\bibitem[{Xia et~al.(2024)Xia, Xing, Du, Yang, Feng, Xu, Yin, and Xiong}]{xia2024fofo}
Congying Xia, Chen Xing, Jiangshu Du, Xinyi Yang, Yihao Feng, Ran Xu, Wenpeng Yin, and Caiming Xiong. 2024.
\newblock Fofo: A benchmark to evaluate llms' format-following capability.
\newblock \emph{arXiv preprint arXiv:2402.18667}.

\bibitem[{Xu et~al.(2023)Xu, Sun, Zheng, Geng, Zhao, Feng, Tao, Lin, and Jiang}]{xu2023wizardlm}
Can Xu, Qingfeng Sun, Kai Zheng, Xiubo Geng, Pu~Zhao, Jiazhan Feng, Chongyang Tao, Qingwei Lin, and Daxin Jiang. 2023.
\newblock Wizardlm: Empowering large pre-trained language models to follow complex instructions.
\newblock In \emph{The Twelfth International Conference on Learning Representations}.

\bibitem[{Yizhi et~al.(2024)Yizhi, Zhang, Qu, Li, Li, Wang, Li, Yuan, Ma, Zhang, Zhou, Liang, Zhang, Ma, Zhang, Li, Huang, Lin, Chen, and Fu}]{yizhi2024cifbench}
Li~Yizhi, Ge~Zhang, Xingwei Qu, Jiali Li, Zhaoqun Li, Zekun Wang, Hao Li, Ruibin Yuan, Yi~Ma, Kai Zhang, Wangchunshu Zhou, Yiming Liang, Lei Zhang, Lei Ma, Jiajun Zhang, Zuowen Li, Stephen~W. Huang, Chenghua Lin, Wenhu Chen, and Jie Fu. 2024.
\newblock Cif-{Bench}: A {Chinese} {Instruction}-{Following} {Benchmark} for {Evaluating} the {Generalizability} of {Large} {Language} {Models}.
\newblock \emph{arXiv}.

\bibitem[{Zhang et~al.(2023)Zhang, Dong, Li, Zhang, Sun, Wang, Li, Hu, Zhang, Wu et~al.}]{zhang2023instruction}
Shengyu Zhang, Linfeng Dong, Xiaoya Li, Sen Zhang, Xiaofei Sun, Shuhe Wang, Jiwei Li, Runyi Hu, Tianwei Zhang, Fei Wu, et~al. 2023.
\newblock Instruction tuning for large language models: A survey.
\newblock \emph{arXiv preprint arXiv:2308.10792}.

\bibitem[{Zheng et~al.(2024)Zheng, Zhang, Zhang, Ye, Luo, and Ma}]{zheng2024llamafactory}
Yaowei Zheng, Richong Zhang, Junhao Zhang, Yanhan Ye, Zheyan Luo, and Yongqiang Ma. 2024.
\newblock \href {http://arxiv.org/abs/2403.13372} {Llamafactory: Unified efficient fine-tuning of 100+ language models}.
\newblock \emph{arXiv preprint arXiv:2403.13372}.

\bibitem[{Zhou et~al.(2023{\natexlab{a}})Zhou, Liu, Xu, Iyer, Sun, Mao, Ma, Efrat, Yu, Yu, Zhang, Ghosh, Lewis, Zettlemoyer, and Levy}]{zhou2023lima}
Chunting Zhou, Pengfei Liu, Puxin Xu, Srini Iyer, Jiao Sun, Yuning Mao, Xuezhe Ma, Avia Efrat, Ping Yu, L.~Yu, Susan Zhang, Gargi Ghosh, M.~Lewis, Luke Zettlemoyer, and Omer Levy. 2023{\natexlab{a}}.
\newblock Lima: Less {Is} {More} for {Alignment}.
\newblock In \emph{Thirty-seventh {Conference} on {Neural} {Information} {Processing} {Systems}}, volume abs/2305.11206.

\bibitem[{Zhou et~al.(2023{\natexlab{b}})Zhou, Lu, Mishra, Brahma, Basu, Luan, Zhou, and Hou}]{ifeval}
Jeffrey Zhou, Tianjian Lu, Swaroop Mishra, Siddhartha Brahma, Sujoy Basu, Yi~Luan, Denny Zhou, and Le~Hou. 2023{\natexlab{b}}.
\newblock Instruction-{Following} {Evaluation} for {Large} {Language} {Models}.
\newblock \emph{arXiv}, abs/2311.07911.

\end{thebibliography}

\appendix

\section{Appendix}
\label{sec:appendix}

\subsection{Case Study on Meta Constraint}
Figure~\ref{fig: Example} shows an example on the ``Limited Text'' meta constraint, which aims at controlling the length of the generated response and ensuring it will not exceed the specified word limit.  In this case, a simple Python function is used to count the words and verify if the response adheres to the word limit defined by the variable `VAR1'. 
Figure~\ref{fig: example2} shows an example on the ``Limited Structure'' meta constraint, which enforces a specific format in the generated output, requiring it to be formatted in the JSON format.

\begin{figure*}[t!]
\centering
\begin{tcolorbox}[colback=gray!00,
                  colframe=black,
                  width=16cm,
                  arc=1.5mm, auto outer arc,
                  left=0.9mm, right=0.9mm,
                  boxrule=0.9pt, colbacktitle = black!65!black
                 ]
\ding{228} ``Category'': ``Limited Text''

\ding{228} ``Level'': 1

\ding{228}  ``Vars'': [{``name'': ``VAR1'', ``type'': ``int'', ``values'': [30, 50, 100]}]

\ding{228} ``Explaination'': ``Use a word count function to verify the response does not exceed the specified limit of VAR1 words.''

\ding{228} ``Python Script''
\begin{verbatim}
def verify_response_limit(response_text, vars, type=0):
    word_limit = int(vars[0])
    word_count = len(response_text.split())
    meets_criteria = word_count <= word_limit
    if type==0:
        return meets_criteria
    else:
        if meets_criteria:
            return 1
        else:
            return 1-(word_count-word_limit)/word_limit
\end{verbatim}

\end{tcolorbox}
\caption{Example of a Limited Text category meta constraint.}
\label{fig: Example}
\end{figure*}

\begin{figure*}[t!]
\centering
\begin{tcolorbox}[colback=gray!00,
                  colframe=black,
                  width=16cm,
                  arc=1.5mm, auto outer arc,
                  left=0.9mm, right=0.9mm,
                  boxrule=0.9pt, colbacktitle = black!65!black
                 ]

\ding{228} ``Category'': ``Limited Structure''

\ding{228} ``Level'': 1

\ding{228} ``Vars'': []

\ding{228} ``Explaination'': ``Entire output should be wrapped in JSON format.''

\ding{228} ``Python Script''
\begin{verbatim}
def verify_json_format(response_text, vars, type=0):
    try:
        response_text=fr'''{response_text}'''
        json_object = json.loads(response_text)
    except ValueError:
        return False
    return True
\end{verbatim}

\end{tcolorbox}
\caption{Example of Limited Structure category in our proposed Meta Constraint.}
\label{fig: example2}
\end{figure*}

\subsection{Case Study on Format Control Instruction}
The format control instructions are instantiated from the meta constraints and general instructions of Alpaca. We add an increasing number of constraints (up to 3) to form instructions of increasing complexity (levels 1 to 3). 
Figure~\ref{fig: 1-level} illustrates an instruction of generating a 3D house model under a single format constraint. The instruction requires the output to be presented in one paragraph and limited to five sentences.
Figure~\ref{fig: 2-level} shows a level-2 instruction constraint where the task is to generate three verbs synonymous with ``to apologize,'' and the response must consist of exactly three sentences. 
Figure~\ref{fig: 3-level} presents an instruction involving three constraints: sentence starting with a specific letter, the response should be exactly 6 paragraphs, and the response must contain between 50 and 110 words. 
Initially, the LLM often struggles to meet all output format constraints, as demonstrated in the failed responses. After fine-tuning, the LLM improves its format control generations and can properly respond to the constraints. 
It is noteworthy that sometimes the format constraints may have conflicts with the original instruction, which may result in negative effects for fine-tuning the model.

\begin{figure*}[ht]
\centering
\begin{tcolorbox}[colback=gray!00,
                  colframe=black,
                  width=16cm,
                  arc=1.5mm, auto outer arc,
                  left=0.9mm, right=0.9mm,
                  boxrule=0.9pt, colbacktitle = black!65!black
                 ]

\textbf{Instruction} \\
Render a 3D model of a house. \\
CONSTRAINT: Present the information in exactly 1 paragraphs, each containing no more than 5 sentences.

\textbf{Successful Response} \\
A 3D model of a house can be created using computer-aided design (CAD) software. The model starts with a foundation, which is the base of the house. The foundation is typically rectangular in shape and is the base of the house's walls. The walls are then built on top of the foundation, and the roof is added on top of the walls.

\textbf{Failed Response} \\
Here is a 3D model of a house: \\
The house is a modern, single-story structure with a rectangular shape. It has a flat roof with a slight slope to allow for rainwater runoff. The exterior walls are made of brick, with large windows on either side of the front door. The door is a simple, rectangular shape with a small awning above it. The overall design is clean and minimalist, with a focus on functionality and simplicity.\\
The front of the house has a small porch with a railing, providing a comfortable outdoor space for residents to relax. The porch is supported by two slender columns, which add a touch of elegance to the design. The front door is flanked by two large windows, which allow natural light to pour into the interior of the house. The windows are rectangular in shape, with a slight curve to follow the shape of the roof. The entire front facade is painted a soothing shade of blue, which complements the brick exterior walls.\\
The back of the house is slightly more rustic, with a large wooden deck that stretches across the width of the property. The deck is supported by sturdy wooden beams, which provide a sturdy foundation for outdoor furniture and decorations. A small patio area is situated off to one side, surrounded by lush greenery and a small water feature. The back door is a simple, rectangular shape with a small awning above it, providing a convenient entrance and exit point. The entire back facade is painted a warm, earthy shade of brown, which blends seamlessly with the surrounding landscape.

\end{tcolorbox}
\caption{Example of level-1 instruction with single format constraint.}
\label{fig: 1-level}
\end{figure*}

\begin{figure*}[t!]
\centering
\begin{tcolorbox}[colback=gray!00,
                  colframe=black,
                  width=16cm,
                  arc=1.5mm, auto outer arc,
                  left=0.9mm, right=0.9mm,
                  boxrule=0.9pt, colbacktitle = black!65!black
                 ]

\textbf{Instruction} \\
Generate three verbs that mean the same as ``to apologize'' \\
CONSTRAINT: Your answer must be composed of exactly 3 sentences. \\
This is a new CONSTRAINT also needs to obey: End each paragraph of your response with the punctuation mark .

\textbf{Successful Response} \\
Exceeding the constraint, I present to you three verbs that mean the same as ``to apologize'': to regret, to excuse, and to make amends. \\ 
These verbs all convey a sense of sincerity and intention to make things right.
\\ To apologize, you can use one of these verbs.

\textbf{Failed Response} \\
Explain: I understand that the response must be composed of exactly three sentences, and that each paragraph must end with the punctuation mark.\\
Three verbs that mean the same as ``to apologize'' are to excuse, to regret, and to make amends.

\end{tcolorbox}
\caption{Example of level-2 instruction with two different format constraints.}
\label{fig: 2-level}
\end{figure*}

\begin{figure*}[t!]
\centering
\begin{tcolorbox}[colback=gray!00,
                  colframe=black,
                  width=16cm,
                  arc=1.5mm, auto outer arc,
                  left=0.9mm, right=0.9mm,
                  boxrule=0.9pt, colbacktitle = black!65!black
                 ]

\textbf{Instruction} \\
List 5 questions for a job interview.\\
Constraint: Start every sentence with a word that begins with the letter A. \\
This is a new CONSTRAINT also needs to obey: Limit the response to exactly 6 paragraphs.\\
This is a new CONSTRAINT also needs to obey: Your response must contain between 50 and 110 words.\\
\textbf{Successful Response} \\
Are you prepared to discuss your qualifications and experience?\\
Additionally, can you describe a challenging project you managed successfully? \\
Analyzing your problem-solving abilities is important to us. \\
Also, what attracted you to this position and our company? \\
Answering this will give insight into your motivation. Are there any questions you have for us about the job or team?\\
Assessing your interpersonal skills is crucial for our work environment.\\
\textbf{Failed Response} \\
Amazingly, the job interview process is an opportunity to showcase skills and experience.\\ 
Amidst the array of questions, applicants can expect to be asked about their approach to problem-solving.\\
Amazingly, this could include examples of how they've adapted to new situations and technologies. \\
Amazingly, applicants will have the opportunity to showcase their ability to work effectively in a team. \\
Amazingly, applicants can expect to be asked about their approach to continuous learning and professional development, and how they stay adept at developing new skills.\\
Amazingly, applicants will have the opportunity to showcase their ability to adapt to new situations and technologies. \\
Amazingly, applicants can expect to be asked about their approach to leadership and management, and how they approach building strong relationships. \\
Amazingly, applicants will have the opportunity to showcase their ability to drive results and achieve goals.

\end{tcolorbox}
\caption{Example of level-3 instruction with three different format constraints.}
\label{fig: 3-level}
\end{figure*}

\end{document}